\documentclass{bmvc2k}

\usepackage{framed,multirow}

\usepackage{amssymb}
\usepackage{latexsym}
\usepackage{graphicx}
\usepackage{subfigure}
\usepackage{multirow}

\usepackage{floatrow}
\newfloatcommand{capbtabbox}{table}[][\FBwidth]

\title{WebCaricature: a benchmark for caricature recognition}

\addauthor{Jing Huo}{huojing@nju.eu.cn}{1}
\addauthor{Wenbin Li}{liwenbin.nju@gmail.com}{1}
\addauthor{Yinghuan Shi}{syh@nju.edu.cn}{1}
\addauthor{Yang Gao}{gaoy@nju.edu.cn}{1}
\addauthor{Hujun Yin}{hujun.yin@manchester.ac.uk}{2}

\addinstitution{
State Key Laboratory for Novel Software Technology\\
Nanjing University\\
Nanjing, China
}
\addinstitution{
School of Electrical and Electronic Engineering\\
The University of Manchester\\
Manchester M13 9PL, U.K.
}

\runninghead{Huo \etal}{WebCaricature: a benchmark for caricature recognition}

\def\etal{\emph{et al}\bmvaOneDot}

\begin{document}

\maketitle

\begin{abstract}
Studying caricature recognition is fundamentally important to understanding of face perception. However, little research has been conducted in the computer vision community, largely due to the shortage of suitable datasets. In this paper, a new caricature dataset is built, with the objective to facilitate research in caricature recognition. All the caricatures and face images were collected from the Web. Compared with two existing datasets, this dataset is much more challenging, with a much greater number of available images, artistic styles and larger intra-personal variations. Evaluation protocols are also offered together with their baseline performances on the dataset to allow fair comparisons. Besides, a framework for caricature face recognition is presented to make a thorough analyze of the challenges of caricature recognition. By analyzing the challenges, the goal is to show problems that worth to be further investigated. Additionally, based on the evaluation protocols and the framework, baseline performances of various state-of-the-art algorithms are provided. A conclusion is that there is still a large space for performance improvement and the analyzed problems still need further investigation.
\end{abstract}

\section{Introduction}
\label{Sec:Introduction}
In the past few years, face recognition performance, even in unconstrained environments, has improved substantially. For instance, the recognition accuracy on the LFW dataset has reached 99\%, even outperforming most humans \cite{sun2014deep}. However, caricature recognition performances by computers are still low \cite{huo2017variation, klare2012towards}.

Caricatures are facial drawings by artists with exaggerations of certain facial parts or features. The exaggerations are often beyond realism and yet the caricatures are effortlessly recognizable by humans. In some cases, people even find that caricatures with exaggerations are easier to recognise as compared to face photos \cite{mauro1992caricature, Perkins}. However, this is not the case for computers. For computers, the basic goal of caricature recognition consists of deciding whether a caricature image and a photo are from the same person. As caricatures are of varying styles and different shape exaggerations, such recognition is difficult.

Caricatures have long fascinated psychologists, neuroscientists and now computer scientists and engineers in their seemingly and grossly distorted views of veridical faces while still possessing distinctively recognizable features of the subjects \cite{mauro1992caricature, sinha2006face, rodriguez2011reverse}. Studying caricatures can offer valuable insights to how face recognition is robustly performed by humans. Studying caricature recognition can also lead to a better understanding of human perception of faces. For example, studies on caricature recognition \cite{mauro1992caricature} have shown that faces may be encoded as distinctive features deviated from prototype faces in human brain. It is thus interesting to see what will happen if this result is used in the current deep learning methods to explicitly encode face features as their distinctiveness from the prototypes. However, this has been rarely studied. Therefore, it is fundamentally imperative to study caricature recognition to shed light on the intrinsic nature of human perception of faces. Computer scientists can take insights gained from psychological studies to develop machine learning methods to further improve caricature and face recognition performance.

\begin{table*}
\newcommand{\tabincell}[2]{\begin{tabular}{@{}#1@{}}#2\end{tabular}}
\caption{Summary of existing datasets relating to caricature recognition.}
\label{Table:SummaryOfCaricatureRecognition}
\footnotesize
\centering
\begin{tabular}{|l|l|l|l|l|}
\hline
& \multicolumn{3}{c|}{Dataset information }\\
\cline{2-4}
& \multicolumn{1}{c|}{Number of subjects} & \multicolumn{1}{c|}{Number of images}  & {\tabincell{l}{Dataset purpose}} \\
\hline
Klare \etal \cite{klare2012towards} & 196 subjects&392 images & \tabincell{l}{Caricature recognition} \\
\hline
Abaci and Akgul \cite{abaci2015matching} & 200 subjects&400 images & \tabincell{l}{Caricature recognition}  \\
\hline
Crowley \etal \cite{BMVC2015_65} & \tabincell{l}{Dev: 1,088 subjects\\NPG: 188 subjects\\Train: 496 subjects} & \tabincell{l}{Dev: 8,528 images\\NPG: 3,128 images\\Train: 257,000 images}  & \tabincell{l}{Painting retrieval} \\
\hline
Mishra \cite{MishraECCV16} & 100 subjects & \tabincell{l}{Caricature: 8,928 images\\Face: 1,000 images} & Cartoon recognition\\
\hline
This paper & 252 subjects & 12,016 images & \tabincell{l}{Caricature recognition}  \\
\hline
\end{tabular}\\
\end{table*}

Currently, there is only limited work on caricature recognition. Besides, there also lack suitable caricature datasets for such research. In fact, there are only four publicly available datasets \cite{klare2012towards, abaci2015matching, MishraECCV16, BMVC2015_65} that are related to caricature recognition, shown in Table \ref{Table:SummaryOfCaricatureRecognition}. The dataset of Klare \etal \cite{klare2012towards} only has 392 images, while the dataset of Abaci and Akgul \cite{abaci2015matching} has 400 images, which certainly limits the study of caricature recognition. Another limitation of these two datasets is that they do not provide benchmark evaluation protocols, which may prove difficult to compare studies and performances. The dataset of Mishra \etal \cite{MishraECCV16} has more images, but only from 100 subjects. Another related dataset is from Crowley \etal\cite{BMVC2015_65}. However, this dataset was developed for painting retrieval rather than caricature recognition.

\begin{figure}
  \centering
  \includegraphics[width=0.99\textwidth]{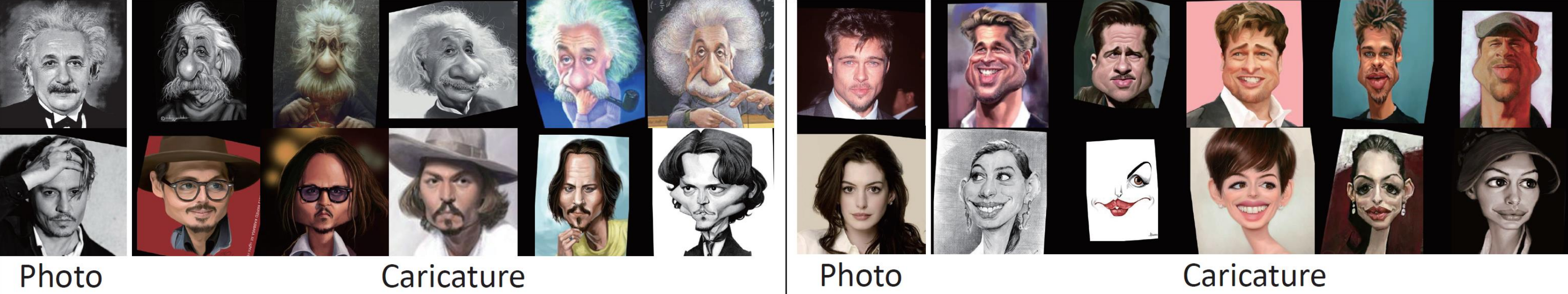}
  \caption{Examples of photos and caricatures. For each person, a photo is given in the first column. The next five columns are corresponding caricatures of various artistic styles, exhibiting large intra-class variability.}\label{Fig:ExampleCaricatures}
\end{figure}

The main contribution of this paper is that we have built a new, larger dataset of caricatures and photos (called WebCaricature\footnote{https://cs.nju.edu.cn/rl/WebCaricature.htm}) with over 12,000 images to facilitate the study on caricature recognition. Examples of photos and caricatures of this new dataset are shown in Fig. \ref{Fig:ExampleCaricatures}. This dataset is much larger than any existing ones and contains more challenging intra-class variations for both photos and caricatures. In addition, we have provided facial landmark annotations on each images of the entire dataset, as well as several evaluation protocols and their baseline performances for comparison.

On caricature recognition, Klare \etal \cite{klare2012towards} proposed to use attribute features, labeled by human via Amazon's Mechanical Turk service. Logistic regression (LR) \cite{freedman2009statistical}, multiple kernel learning (MKL) \cite{bach2008consistency}, and support vector machines (SVM) \cite{cortes1995support} were then used to calculate the similarity of a caricature and a photo based on these features. Authors in \cite{ouyang2014cross} proposed to learn a facial attribute model. Facial attribute features were then combined with low-level features for recognition using canonical correlation analysis (CCA). Abaci and Akgul \cite{abaci2015matching} proposed a method to extract facial attribute features for photos. For caricatures, the attribute features were manually labeled. Then the weights of these attribute features were learned by a genetic algorithm (GA) or LR and then used for recognition.

The existing work is monotonous with main effort on extracting facial attribute features. Little work \cite{huo2017variation} has discussed the possibility of applying low-level features and/or deep learning features for caricature recognition. Besides, the existing work did not consider the detection and facial landmark localization of caricatures, an important process for automatic recognition. To complement this absence of related work, a caricature recognition framework is presented. Similar to the current face recognition framework, we present challenges of caricature recognition within the framework as compared to face recognition. Based on this newly constructed dataset and its evaluation protocols, a combination of many state-of-the-art methods under the framework are tested. The experiments not only help to demonstrate the challenges but also offer baseline performances. With the presented framework, future researchers are advocated to focus more on key challenges rather than stalling at attribute feature learning. To summarize, in additional to a new large caricature dataset, WebCaricature, this paper presents a framework for caricature recognition. With the framework, we analyze the challenges of caricature recognition and hence provide baseline performances on the dataset.

\section{Dataset Collection}
\label{Sec:DatasetCollection}

\subsection{Collection Process} For collecting both face photos and caricature drawings, we first drew up a list of names of celebrities. For each person, we manually searched for caricatures and photos and saved these images. All the photos were searched using the Google image search, while caricatures were mainly from the Google image search and the Pinterest website.

After all the images were downloaded, a program was written to detect and remove duplicated caricatures and photos. This was done by extracting features of any two caricatures or photos, pairs of features with large similarities were selected. We then manually checked whether they were indeed duplicated images. The process resulted in a dataset of 252 subjects with a total of 6042 caricatures and 5974 photos. For each person, the number of caricatures ranges from 1 to 114 and the number of photos from 7 to 59.

\subsection{Labeling Process} For each caricature, 17 landmarks were manually labeled. The first four landmarks are the basic face contours. The next four are eye brows. Landmarks 9-12 are eye corners. The 13th landmark is nose tip. The 14th-17th are mouth contours. Details can be found from the dataset website.

The landmark labeling procedure for photos was different. For each photo, firstly we used the facial landmark detection software provided by the Face++ software \cite{face++}. The software could locate up to 83 facial landmarks, from which we used only the parts of the face landmarks that corresponded to the 2nd-17th landmarks labelled on the caricatures. The first landmark was then manually labeled for each of the photos. Note that there were some photos the software did not label well or failed to locate landmarks. For those cases, we manually corrected the labeling or manually labeled them with the same scheme as for the caricatures. After the above two processes, 17 landmarks for each caricature and photo were obtained.

\section{Evaluation Protocols}
\label{Sec:Protocols}
Generally, face recognition can be categorized into two categories: verification and identification. As the two kinds of recognition generally deal with different problems. To promote the study of both caricature verification and identification, four experimental protocols covering both scenarios are developed on the constructed WebCaricature dataset. All the protocols are public available along with the dataset.

\textbf{Caricature Verification.} The task of caricature verification is to verify whether a caricature and a photo are from the same person. Therefore, the algorithm is presented with a pair of images (a caricature and a photo) and the output is either yes or no. To evaluate the performance of this protocol, we have built two settings, similar to that used for LFW \cite{LFWTech}. One is image restricted setting and the other is image unrestricted setting. For each setting, the dataset is divided into training and testing sets.

In the image restricted setting, there are two views. View 1 is for parameter tuning and View 2 for testing. Pairs of images are provided with the information on whether the pairs are from the same person. There is no extra identity information. For training, only the provided pairs should be used. In the image unrestricted setting, there are also two views. In this setting, training images together with person identities are given. Therefore, as many pairs as possible can be formulated for training. For View 1 of both restricted and unrestricted settings, the proportion of subjects in training and testing were set to close to 9:1. For View 2, 10-fold cross validation should be used. Therefore, to report results on View 2, training must be carried out for ten times; and each time, 9 folds of the data are used for training and the remaining fold for testing. For these two settings, researchers are encouraged to use the receiver operating characteristic (ROC) curve, area under the ROC curve (AUC), VR@FAR=0.1\%, and VR@FAR=1\% to report performance, where VR@FAR=0.1\% corresponds to the verification rate (VR) with false accept rate (FAR) equal to 0.1\%, and VR@FAR=1\% denotes VR with FAR of 1\%.

\textbf{Caricature Identification.}
The task of caricature identification can be formulated into two settings. One is to find the corresponding photo(s) of a given caricature from a photo gallery (denoted as Caricature to Photo or C2P). The second is to find the corresponding caricature(s) for a given photo from a set of caricatures (denoted as Photo to Caricature or P2C). For each of the two settings, there are two views. View 1 is for development and parameter tuning and View 2 for testing. While generating data for these two views, subjects were evenly split to training and testing sets. For View 2, the dataset was randomly split for ten times. Thus for results reporting, the algorithm should run ten times and report the average results. Besides, for C2P, photos are used as the gallery and caricatures are used as the probe. For each subject in the gallery set, only one photo is randomly selected. Reversely, for P2C, caricatures are used as the gallery and each subject has only one randomly selected caricature in the gallery. For these two settings, we encourage researchers to use the cumulative match curve (CMC) for evaluation and report average Rank-1 and Rank-10 results.

\section{Recognition Framework and Challenges}
\label{Sec:Framework}
\begin{figure*}
  \centering
  \includegraphics[scale = 0.34]{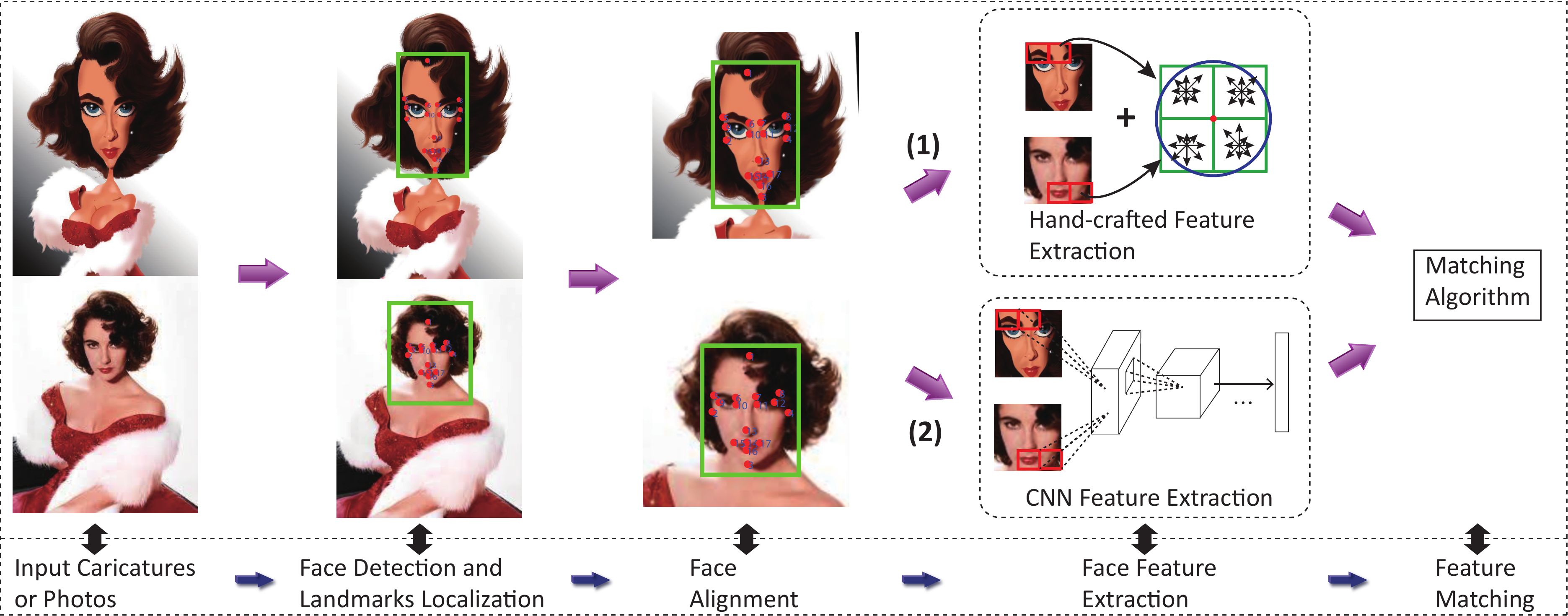}\\
  \caption{Caricature recognition framework. The first step is face detection and landmark localization. Then faces are aligned. With aligned face images, features are extracted. As illustrated, two kinds of feature extraction methods can be used, hand-crafted or convolutional neural network based. The last step is to measure the similarity of extracted features.}\label{Fig:RecognitionProcess}
\end{figure*}

As reviewed in Section \ref{Sec:Introduction}, most existing work on caricature recognition tries to extract descriptive attribute features for recognition. In this paper, we illustrate how to solve the recognition problem without using attributes, as they are subjective and require extensive manual labeling. The proposed recognition framework is shown in Fig. \ref{Fig:RecognitionProcess}. There are four main steps in the framework. The process is similar to the traditional face recognition process, albeit more challenges at each step.

\textbf{Caricature Face Detection and Landmark Localization.} The first step of the recognition process is to perform caricature face detection and landmark localization. For the detection, as caricatures can vary greatly in artistic styles and facial appearances, it is more difficult to find detection patterns of caricatures compared to real faces in photos. For localizing landmarks, the shapes of caricatures and the positions of eyes, noses and mouths can be exaggerated and sometimes appear at odd positions beyond realism, making most of the existing localization techniques unusable. Although this paper is devoted mainly for studying caricature recognition, further research on automatic caricature detection and landmark localization can also be performed on the proposed dataset, as our dataset only provides 17 landmarks for each caricature.

\begin{figure}
  \centering
  \includegraphics[scale = 0.35]{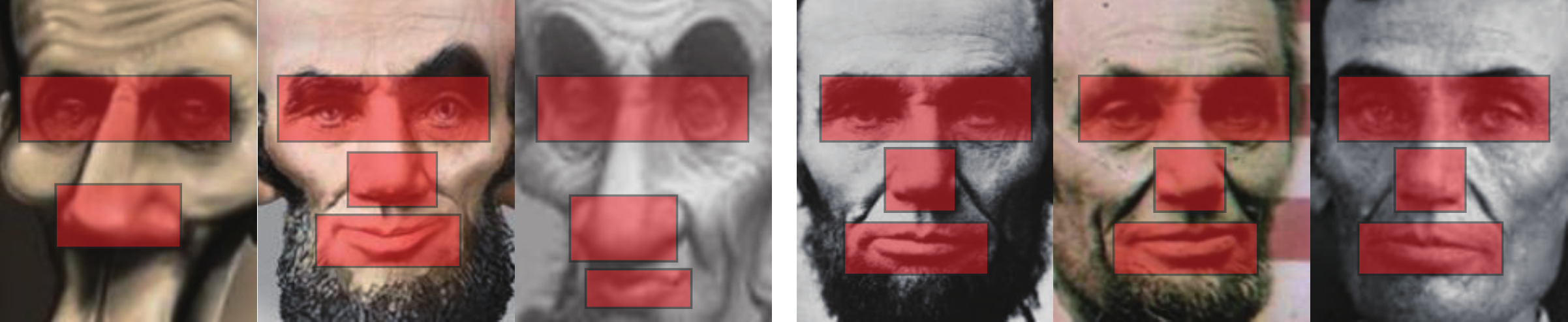}\\
  \caption{Illustration of the misalignment problem of caricatures. Nose and mouth parts appear at different positions on the aligned caricature images (first three). Photos are fairly well aligned (last three).}\label{Fig:Misalign}
\end{figure}

\textbf{Face Alignment.} The second step is face alignment. For this task, the objective is to make the eyes, noses or mouths appear at the same positions on two cropped face images, such that the two images are comparable. Three alignment methods (details can be found in Section \ref{SubSec:Verification_Align}), which work well for traditional face recognition, were tested in this paper. However, we found that in caricature recognition, caricatures can still be misaligned even after alignment is applied, because many facial parts are exaggerated and distorted. An illustration of misaligned caricatures after applying the eye-based alignment method is given in Fig. \ref{Fig:Misalign}. The eye-based alignment method forces the eye distance to a fixed number and the center of two eyes at a fixed position in an aligned face image. As can be seen, the photos are well aligned. Eyes of the caricatures are also well aligned. However, there are mismatches in nose and mouth areas of the caricatures. Hence for caricature recognition, more sophisticated methods need to be developed to address the misalignment problem.

\textbf{Face Feature Extraction.} The goal of face feature extraction is to extract robust and discriminant person specific features. Besides the challenges of traditional face feature extraction, such as illumination, expression and viewing angle variations, one major challenge for caricature feature extraction is to remove the variation across different drawing styles (also identified as modality variation, as photos and caricatures can be seen as from two modalities). Besides, if misalignment problem is not addressed well in the previous step, the feature extraction step should also take the exaggeration and distortion problems into consideration, making the feature extraction even more difficult.

In the experiments, we have tested two kinds of feature extraction methods. The first kind is hand-crafted feature extraction: local binary patterns (LBP) \cite{1717463}, Gabor \cite{999679} and scale invariant feature transform (SIFT) \cite{Lowe2004} features were tested. The second is the convolutional neural network (CNN) based method, where the VGG-Face model \cite{Parkhi15} was directly adopted. In these experiments, the feature extraction process was the same for caricatures and photos. However, researchers are advocated to design more sophisticated feature extraction methods to solve both the modality variation and misalignment problems.

\textbf{Matching Algorithm.} The last step is to compute the similarity of two sets of features from a caricature and a photo to decide whether they are from the same person. If the misalignment problem is not addressed and the image style variation is not removed, then the matching algorithm should consider removing both alignment and modality variations first. In this case, cross-modality metric learning methods can be useful \cite{huo2017variation, 7959077, 8246530}. Otherwise, if the two problems are fairly well addressed by the previous steps, traditional matching algorithm can be directly applied.

For this step, nine subspace and metric learning methods were tested, including both single modality and cross-modality based methods. For single modality based methods, principal component analysis (PCA) \cite{kim1996face}, kernel discriminant analysis (KDA) \cite{Cai11SRKDA}, keep it simple and straightforward metric (KISSME) \cite{koestinger2012large}, the information-theoretic metric learning (ITML) \cite{davis2007information}, the large margin nearest neighbor (LMNN) \cite{weinberger2009distance} were tested. For cross-modality based methods, tested methods include canonical correlation analysis (CCA) \cite{hotelling1936relations}, multi-view discriminant analysis (MvDA) \cite{kan2016multi}, coupled spectral regression (CSR) \cite{lei2009coupled}, kernel coupled spectral regression (KCSR) \cite{lei2009coupled}. In the experiments, we demonstrate that removing modality variation by cross-modality based methods, such as CSR and KCSR, can largely improve the performance.

\section{Baseline Performance}
\label{Sec:Baseline}
Following the proposed evaluation protocols and the caricature recognition framework, various alignment methods, feature extraction methods, subspace and metric learning methods were tested on the developed dataset to demonstrate the challenges of caricature recognition and also to provide baseline performances on the WebCaricature dataset. For the following experiments, under the verification settings, adopted evaluation performance measures include VR@FAR=0.1\% (denoted as FAR=0.1\% in Tables \ref{Table:AlignmentVerification}, \ref{Table:SubspaceVerification}, \ref{Table:Best_Results_Verification}), VR@FAR=1\% (denoted as FAR=1\% in Tables \ref{Table:AlignmentVerification}, \ref{Table:SubspaceVerification}, \ref{Table:Best_Results_Verification}) and AUC. Under the identification settings, the evaluation performance measures used include Rank-1 and Rank-10.

\subsection{Influence of Alignment Methods}
\label{SubSec:Verification_Align}
To study the influence of alignment, three alignment methods were tested. The alignment process of the first method was to rotate the image first to align the two eyes to horizontal position. Then, the image was resized to make eye distance of 75 pixels. After this step, the cropped region was defined by making the eye center to the upside of the bounding box by 70 pixels and to the left/right side of the bounding box by 80 pixels. The bounding box is of size $160\times 200$. This alignment method is denoted as eye location based. In the second alignment method, the first two steps were the same. Then, small regions of $40\times 40$ centered at each of the 17 labeled landmarks were also cropped out. This alignment method is denoted as landmark based. The third alignment method was done according to the face contour (defined by the landmarks 1-4). Then the bounding box was enlarged by a scale of 1.2 in both width and height. With the enlarged bounding box, the face image inside the bounding box was resized to $160\times 200$. This alignment method is denoted as bounding box based. The three alignment methods were combined with two feature extraction methods (SIFT and VGG-Face). The resulting feature vectors were combined with PCA for evaluation and all the principal components were kept.

\begin{table*}
\caption{Results of various alignment (1-5 rows) and feature extraction methods (6-9 rows) under restricted and unrestricted settings. }
\label{Table:AlignmentVerification}
\centering
\scriptsize
\begin{tabular}{|l|l|l|l||l|l|l|}
\hline
\multirow{2}*{Method}&\multicolumn{3}{c||}{Restricted}&\multicolumn{3}{c|}{Unrestricted}\\
\cline{2-7}
 & FAR=0.1\% (\%)  & FAR=1\% (\%)  & AUC & FAR=0.1\% (\%)  & FAR=1\% (\%)  & AUC  \\
 \hline
SIFT-Eye & $2.78\pm0.46$ & $10.46\pm1.43$ & $0.738\pm 0.016$ & $2.74\pm 0.75$ & $10.74\pm 1.53$ & $0.734\pm 0.016$  \\
\hline
SIFT-Land & $\mathbf{4.67}\pm\mathbf{1.08}$ & $\mathbf{15.39}\pm\mathbf{1.85}$ & $\mathbf{0.777}\pm \mathbf{0.017}$ & $\mathbf{4.43}\pm \mathbf{0.82}$ & $\mathbf{15.24}\pm \mathbf{2.03}$ & $\mathbf{0.780}\pm \mathbf{0.017}$ \\
\hline
SIFT-Box & $2.96\pm0.72$ & $9.15\pm1.17$ & $0.720\pm 0.018$ & $3.42\pm 0.52$ & $11.76\pm 1.21$ & $0.724\pm 0.013$  \\
\hline
VGG-Eye  & $21.42\pm2.02$ & $40.28\pm2.91$ & $0.896\pm 0.013$ & $19.24\pm 1.95$ & $40.88\pm 2.23$ & $0.898\pm 0.007$ \\
\hline
VGG-Box & $\mathbf{28.42}\pm\mathbf{2.04}$ & $\mathbf{55.53}\pm\mathbf{2.76}$ & $\mathbf{0.946}\pm \mathbf{0.009}$ & $\mathbf{32.07}\pm \mathbf{2.60}$ & $\mathbf{56.76}\pm \mathbf{2.35}$ & $\mathbf{0.946}\pm \mathbf{0.005}$\\
\hline
\hline
LBP-Eye & $0.33\pm0.15$ & $1.92\pm0.38$ & $0.600\pm 0.015$ & $0.19\pm 0.04$ & $1.65\pm 0.27$ & $0.597\pm 0.006$ \\
\hline
Gabor-Eye & $3.23\pm0.74$ & $9.75\pm1.36$ & $0.716\pm 0.017$ & $2.76\pm 0.47$ & $10.36\pm 1.54$ & $0.718\pm 0.016$ \\
\hline
SIFT-Eye & $2.78\pm0.46$ & $10.46\pm1.43$ & $0.738\pm 0.016$ & $2.74\pm 0.75$ & $10.74\pm 1.53$ & $0.734\pm 0.016$\\
\hline
VGG-Eye & $\mathbf{21.42}\pm\mathbf{2.02}$ & $\mathbf{40.28}\pm\mathbf{2.91}$ & $\mathbf{0.896}\pm \mathbf{0.013}$ & $\mathbf{19.24}\pm \mathbf{1.95}$ & $\mathbf{40.88}\pm \mathbf{2.23}$ & $\mathbf{0.898}\pm \mathbf{0.007}$\\
\hline
\end{tabular}
\end{table*}

The first 5 rows of Table \ref{Table:AlignmentVerification} summarise the results. It can be seen: (1) for SIFT features, it is obvious that landmark based feature extraction (denoted with a suffix `-Land') obtains the best performance. The results of eye location based (denoted by suffix `-Eye') and bounding box based (denoted by suffix `-Box') methods are worse. \textit{In fact, landmark based method is a hard alignment method that forces two patches of images at the same landmark to be directly compared so as to alleviate misalignment problem. Hence it is consistently the best for extracting hand-crafted features, such as SIFT.} (2) For deep features, the bounding box based alignment is much better than the eye location based method. This is perhaps because that the bounding box based alignment can retain more face information. The cropped face images of eye location based method may miss certain part of the face such as chin and mouth due to the exaggerations in caricatures. \textit{The deep learning based methods may have mechanisms to alleviate misalignments. Thus the results of bounding box based alignment are better. Note that landmark based method is not applicable for deep learning, but it is straightforward to think that introducing a hard alignment scheme into deep learning would address the misalignment problem.}

\subsection{Evaluation of Feature Extraction Schemes}
To study the influence of features, LBP, Gabor, SIFT and CNN (VGG-Face) is adopted for experiments, where the first three are hand-crafted features. For all these feature extraction methods, eye location based alignment is used. All the extracted features were combined with PCA for evaluation. In Table \ref{Table:AlignmentVerification}, 6-9 rows summarizes results. From the table, SIFT feature is almost the best among the hand-crafted features. Gabor is the next. The results of LBP are the worst. The results of VGG-Face are the best, significantly improving over that of SIFT. This illustrates the superiority of deep learned features. Note that VGG-Face was not fine-tuned in this application. Still, the performance is much better than that of any hand-crafted features. \textit{However, under the eye location based alignment setting, the results of VGG-Face for VR@FAR=0.1\% and VR@FAR=1\% are 21.42\% and 40.28\% for the restricted setting and 19.24\% and 40.88\% for the unrestricted setting. Thus, there is still much room for improvement even with deep learning.} Note that there are many other modern deep learning based algorithms that can also be used as baseline. Besides, fine-tuning VGG-Face model is also an interesting approach. These results will be updated in our future work.

\begin{table*}
\caption{Results of various learning methods under restricted and unrestricted verification settings.}
\label{Table:SubspaceVerification}
\centering
\scriptsize
\begin{tabular}{|l|l|l|l||l|l|l|}
\hline
\multirow{2}*{Method}&\multicolumn{3}{c||}{Restricted}&\multicolumn{3}{c|}{Unrestricted}\\
\cline{2-7}
& FAR=0.1\% (\%) & FAR=1\% (\%) & AUC & FAR=0.1\% (\%) & FAR=1\% (\%) & AUC \\
\hline
Euc & $2.12\pm0.68$ & $8.28\pm1.13$ & $0.661\pm 0.014$ & $2.56\pm 0.38$ & $8.40\pm 0.87$ & $0.661\pm 0.009$ \\
\hline
PCA & $4.67\pm1.08$ & $15.39\pm1.85$ & $0.777\pm 0.017$ & $4.43\pm 0.82$ & $15.24\pm 2.03$ & $0.780\pm 0.017$ \\
\hline
KDA & - & - & -  & $6.62\pm 1.37$ & $24.23\pm 3.26$ & $0.875\pm 0.014$  \\
\hline
KissME & $4.55\pm1.07$ & $12.15\pm1.73$ & $0.724\pm 0.011$ & $4.56\pm 0.89$ & $14.66\pm 1.70$ & $0.781\pm 0.016$  \\
\hline
ITML & $\mathbf{5.08}\pm\mathbf{1.82}$ & $\mathbf{18.07}\pm\mathbf{4.72}$ & $\mathbf{0.841}\pm \mathbf{0.018}$ & $5.35\pm 1.20$ & $18.48\pm 2.40$ & $0.828\pm 0.016$ \\
\hline
LMNN & - & - & - & $6.59\pm 1.61$ & $21.37\pm 3.22$ & $0.842\pm 0.014$  \\
\hline
CCA & $4.77\pm0.68$ & $12.96\pm1.40$ & $0.775\pm 0.016$ & $5.02\pm 1.19$ & $17.66\pm 2.49$ & $0.812\pm 0.017$ \\
\hline
MvDA & - & - & - & $1.41\pm 0.37$ & $8.29\pm 0.72$ & $0.753\pm 0.014$ \\
\hline
CSR & - & - & -  & $\mathbf{11.76}\pm \mathbf{2.72}$ & $31.86\pm 3.85$ & $0.887\pm 0.013$ \\
\hline
KCSR & - & - & -  & $11.66\pm 2.69$ & $\mathbf{32.00}\pm \mathbf{3.94}$ & $\mathbf{0.888}\pm \mathbf{0.013}$  \\
\hline
\end{tabular}
\end{table*}

\begin{table*}
\caption{Results of various learning methods under C2P and P2C identification settings. }
\label{Table:SubspaceIdentification}
\centering
\scriptsize
\begin{tabular}{|l|l|l||l|l|}
\hline
\multirow{2}*{Method}&\multicolumn{2}{c||}{C2P}&\multicolumn{2}{c|}{P2C}\\
\cline{2-5}
& Rank-1 (\%) & Rank-10 (\%)& Rank-1 (\%) & Rank-10 (\%)\\
\hline
Euc & $13.38\pm 1.10$  & $38.40\pm 1.92$  & $9.04\pm 0.80$  & $29.63\pm 1.35$ \\
\hline
PCA & $15.63\pm 0.82$  & $43.48\pm 1.69$  & $12.47\pm 1.14$  & $40.13\pm 1.53$ \\
\hline
KDA & $19.32\pm 1.36$  & $56.77\pm 1.58$  & $18.92\pm 1.35$  & $57.19\pm 2.61$ \\
\hline
KissME & $15.16\pm 1.63$  & $43.95\pm 2.13$  & $13.30\pm 1.18$  & $43.63\pm 1.64$ \\
\hline
ITML & $15.25\pm 3.07$  & $46.39\pm 6.46$  & $16.48\pm 1.77$  & $49.88\pm 2.29$ \\
\hline
LMNN & $17.92\pm 0.86$  & $50.58\pm 1.72$  & $15.90\pm 1.73$  & $48.08\pm 1.95$ \\
\hline
CCA & $10.84\pm 0.78$  & $40.76\pm 1.08$  & $10.73\pm 0.94$  & $41.12\pm 1.87$ \\
\hline
MvDA & $4.77\pm 0.74$  & $27.73\pm 1.90$  & $4.71\pm 0.87$  & $27.19\pm 2.55$ \\
\hline
CSR & $\mathbf{25.18}\pm \mathbf{1.39}$  & $60.95\pm 1.20$  & $23.36\pm 1.47$  & $60.27\pm 1.97$ \\
\hline
KCSR & $24.87\pm 1.50$  & $\mathbf{61.57}\pm \mathbf{1.37}$  & $\mathbf{23.42}\pm \mathbf{1.57}$  & $\mathbf{60.95}\pm \mathbf{2.34}$ \\
\hline
\end{tabular}
\end{table*}

\subsection{Evaluation of Different Matching Methods}
\label{Sec:EvalLearningVerification}
In this section, we tested several single modality subspace learning methods as mentioned in Section \ref{Sec:Framework}, including PCA \cite{kim1996face}, KDA \cite{Cai11SRKDA} and multi-modality subspace learning methods such as, CCA \cite{hotelling1936relations}, MvDA \cite{kan2016multi}, CSR \cite{lei2009coupled} and KCSR \cite{lei2009coupled}. The state-of-the-art single modality metric learning methods include the KISSME \cite{koestinger2012large}, ITML \cite{davis2007information} and LMNN \cite{weinberger2009distance}. Landmark based alignment and SIFT features were used for comparing all these methods. Prior to applying these methods, PCA was applied.

As the restricted setting only provides information that two images are either of the same class or not and algorithms such as KDA, LMNN, MvDA, CSR, KCSR require explicit label information for each image, they are not applicable for the image restricted setting. From Table \ref{Table:SubspaceVerification}, the best result was achieved by ITML under the restricted setting. For the unrestricted setting, the best and second best results were achieved by CSR and KCSR, respectively. \textit{In summary, all these learning methods were better than simple Euclidean distance on original features.} For the unrestricted setting, the performance of CSR and KCSR was the best, because they were designed for cross-modality subspace learning. \textit{This suggests that further studies on cross-modality metric learning or cross-modality subspace learning will be beneficial, due to limited work on this direction.} Under the identification settings, from the results in Table \ref{Table:SubspaceIdentification}, CSR and KCSR achieved similarly the best results. The best rank-1 performance for C2P setting was only $25.18 \pm 1.39$ and ${23.42}\pm {1.57}$ for P2C setting. \textit{This means that there is a large room for improvement on these two settings with the traditional face recognition process.}

\subsection{Summary of Results}
A summary of the best combinations of the methods at three stages are provided as baselines. Results under verification settings are given in Table \ref{Table:Best_Results_Verification}. For identification settings, the results are summarized in Table \ref{Table:Best_Results_Identification}. From the tables, the deep learning feature based methods outperform the hand-crafted feature based methods to a large extent. Another observation is that the results of VGG-Face can be further improved with KCSR. \textit{This is mainly because that VGG-Face is trained using only photos and may not be able to deal with modality variations. With the help of KCSR to further remove modality variations, the performance can be enhanced. Thus one future direction is to develop end-to-end modality invariant deep learning methods.} Another finding is that performances, \textit{VR@FAR=0.1\%, VR@FAR=1\%, Rank-1 and Rank-10, of deep learning based methods are still far from satisfactory, indicating that there is still room for improvement.} Lastly, although bounding box based alignment is better than eye based alignment, as analyzed in Section \ref{SubSec:Verification_Align}, \textit{there is still lack of good alignment methods of caricatures for deep learning.}

\begin{table*}
\begin{floatrow}
\capbtabbox{
\caption{Summary of results under restricted and unrestricted settings. }
\label{Table:Best_Results_Verification}}
{\centering
\scalebox{0.668}{
\begin{tabular}{|l|l|l|l|}
\hline
\multicolumn{4}{|c|}{Restricted}\\
\hline
Method& FAR=0.1\% (\%)  & FAR=1\% (\%) & AUC \\
\hline
SIFT-Land-ITML & $5.08\pm1.82$ & $18.07\pm4.72$ & $0.841\pm 0.018$\\
\hline
VGG-Eye-PCA & $21.42\pm2.02$ & $40.28\pm2.91$ & $0.896\pm 0.013$\\
\hline
VGG-Eye-ITML & $18.97\pm3.90$ & $41.72\pm5.83$ & $0.911\pm 0.014$\\
\hline
VGG-Box-PCA & $28.42\pm2.04$ & $55.53\pm2.76$ & $0.946\pm 0.009$\\
\hline
VGG-Box-ITML & $\mathbf{34.94}\pm\mathbf{5.06}$ & $\mathbf{57.22}\pm\mathbf{6.50}$ & $\mathbf{0.954}\pm \mathbf{0.010}$\\
\hline
\hline
\multicolumn{4}{|c|}{Unrestricted}\\
\hline
Method& FAR=0.1\% (\%) & FAR=1\% (\%) & AUC \\
\hline
SIFT-Land-KCSR & $11.66\pm2.69$ & $32.00\pm3.94$ & $0.888\pm 0.013$\\
\hline
VGG-Eye-PCA & $19.24\pm1.95$ & $40.88\pm2.23$ & $0.898\pm 0.007$\\
\hline
VGG-Eye-KCSR & $23.46\pm2.65$ & $48.57\pm2.73$ & $0.925\pm 0.007$\\
\hline
VGG-Box-PCA & $32.07\pm2.60$ & $56.76\pm2.35$ & $0.946\pm 0.005$\\
\hline
VGG-Box-KCSR & $\mathbf{39.09}\pm\mathbf{3.62}$ & $\mathbf{65.82}\pm\mathbf{2.48}$ & $\mathbf{0.963}\pm \mathbf{0.004}$\\
\hline
\end{tabular}
}
}
\capbtabbox{
\caption{Summary of results under C2P and P2C settings.}
\label{Table:Best_Results_Identification}}
{
\scalebox{0.668}{
\centering
\begin{tabular}{|l|l|l|}
\hline
\multicolumn{3}{|c|}{C2P}\\
\hline
Method& Rank-1 (\%) & Rank-10 (\%)\\
\hline
SIFT-Land-KCSR & $24.87\pm 1.50$  & $61.57\pm 1.37$ \\
\hline
VGG-Eye-PCA & $35.07\pm 1.84$  & $71.64\pm 1.32$ \\
\hline
VGG-Eye-KCSR & $39.76\pm 1.60$  & $75.38\pm 1.34$ \\
\hline
VGG-Box-PCA & $49.89\pm 1.97$  & $84.21\pm 1.08$ \\
\hline
VGG-Box-KCSR & $\mathbf{55.41}\pm \mathbf{1.41}$  & $\mathbf{87.00}\pm \mathbf{0.92}$ \\
\hline
\hline
\multicolumn{3}{|c|}{P2C}\\
\hline
Method& Rank-1 (\%)  & Rank-10 (\%) \\
\hline
SIFT-Land-KCSR & $23.42\pm 1.57$  & $60.95\pm 2.34$ \\
\hline
VGG-Eye-PCA & $36.18\pm 3.24$  & $68.95\pm 3.25$ \\
\hline
VGG-Eye-KCSR & $40.67\pm 3.61$  & $75.77\pm 2.63$ \\
\hline
VGG-Box-PCA & $50.59\pm 2.37$  & $82.15\pm 1.31$ \\
\hline
VGG-Box-KCSR & $\mathbf{55.53}\pm \mathbf{2.17}$  & $\mathbf{86.86}\pm \mathbf{1.42}$ \\
\hline
\end{tabular}
}
}
\end{floatrow}
\end{table*}

\section{Conclusions}
\label{Sec:Conclusion}
A new benchmark dataset of face caricatures and photos is presented, together with a framework and protocols, to facilitate caricature recognition and tackling its challenges. The main contribution of this paper includes: a large caricature dataset of 252 people with 6024 caricatures and 5974 photos, which is made publicly available. Facial landmarks, evaluation protocols and baseline performances are provided on the dataset. Following these protocols and the framework, a set of face alignment methods, hand-crafted and deep learning features, and various subspace and metric learning methods are tested. A conclusion is that there is still room for improvement even with the best results. With this dataset and from the baseline evaluations, there are several future directions. Caricature face landmark detection is of great interest and a key step for caricature recognition. As the performance on this dataset is still far from saturated, future work on caricature and face feature extraction and cross-modal metric learning methods are also promising directions.

\section*{Acknowlegement} This work was supported in part by the National Science Foundation of China under Grant 61432008, Grant 61673203, in part by the Young Elite Scientists Sponsorship Program by CAST under Grant YESS 2016QNRC001, and in part by the Collaborative Innovation Center of Novel Software Technology and Industrialization.

\bibliography{refs}

\end{document}